\documentclass{article}
\usepackage[preprint, plain]{neurips_2023}

\usepackage[utf8]{inputenc} 
\usepackage[T1]{fontenc}    
\usepackage{hyperref}	    
\usepackage{url}	    
\usepackage{array}
\usepackage{booktabs}	    
\usepackage{amsfonts}	    
\usepackage{nicefrac}	    
\usepackage{microtype}	    
\usepackage{tabu}
\usepackage[table]{xcolor}	    
\usepackage{xspace}
\usepackage{soul}
\usepackage{arydshln}
\usepackage{makecell}
\usepackage{graphicx}
\usepackage{enumitem}
\usepackage{multirow}
\usepackage[symbol*]{footmisc}
\usepackage{geometry} 
\newcolumntype{L}[1]{>{\raggedright\arraybackslash}p{#1}}
\newcolumntype{P}[1]{>{\raggedright\arraybackslash}p{#1}}

\title{Enhancing Retrieval in QA Systems with Derived Feature Association}

\author{%
    Keyush Shah \quad Abhishek Goyal  \quad Isaac Wasserman \\
    Department of Computer and Information Science\\
    University of Pennsylvania\\
    Philadelphia, PA 19104 \\
    \texttt{\{keyush06, abhi2358, isaacrw\}@seas.upenn.edu}
}

\begin{document}
\setcitestyle{numbers}

\maketitle

\newcommand{\methodlongname}{\textbf{R}etrieval from \textbf{AI}
    \textbf{D}erived \textbf{D}ocuments\xspace}
\newcommand{\methodacronym}{RAIDD\xspace}

\definecolor{correct}{RGB}{133, 255, 130}
\definecolor{incorrect}{RGB}{255, 105, 105}
\definecolor{lightgray}{RGB}{210, 210, 220}
\definecolor{offwhite}{RGB}{250, 250, 255}

\newcommand{\highlight}[2]{
    \kern-0.5em
    \sethlcolor{#1} 
    \hl{#2} 
    \kern-0.5em
}
\begin{abstract}
    Retrieval augmented generation (RAG) has become the standard in long-context question answering (QA) systems. However, typical implementations of RAG rely on a rather naive retrieval mechanism, in which texts whose embeddings are most similar to that of the query are deemed most relevant. This has consequences in subjective QA tasks, where the most relevant text may not directly contain the answer. In this work, we propose a novel extension to RAG systems, which we call \methodlongname (\methodacronym). \methodacronym leverages the full power of the LLM in the retrieval process by deriving inferred features, such as summaries and example questions, from the documents at ingest. We demonstrate that this approach significantly improves the performance of RAG systems on long-context QA tasks. \footnote{\label{Github Link}https://github.com/isaacwasserman/LongRAG}

\end{abstract}

\section{Introduction}
\subsection{Preliminaries}
First introduced by \cite{lewis2020retrievalaugmented}, retrieval augmented generation (RAG) allows LLMs to pull relevant information into context from a cache of documents. The system allows these models to access up-to-date information, rely less on their parameterized-memory, and leverage a large corpus of documents during generation, despite their limited context window \cite{zhao2024retrievalaugmented}. RAG extends LLMs with a retrieval mechanism that takes a query, selects the most relevant texts from a given corpus, and hands them to the generator to inform its answer. Early approaches, were optimized end-to-end, using a jointly learned retriever and generator that communicated through a shared embedding space \cite{lewis2020retrievalaugmented}. However, the requirement that such a system must be trained from scratch for each choice of generator architecture makes this approach expensive and cumbersome given the rapid pace with which new LLMs are developed. However, this paradigm was subverted by \cite{shi2023replug}, which assumes the generator to be black-box, training a generator-agnostic retriever that simply prepends the retrieved text to the generator's input. In practice, the retriever is often further simplified to score documents based on their cosine similarity in a pretrained embedding space (dense retrieval); also popular is the use of BM25, a simple term-frequency based similarity metric (sparse retrieval) \cite{zhao2024retrievalaugmented}.

\subsection{Motivation}
RAG systems, especially those which rely on embedding cosine similarity or BM25 to measure relevance, are fast and remarkably effective for answering questions whose answers are explicitly stated in the text. However, from a user's perspective, this is only marginally more effective than a simple \verb^ctrl+f^ search. We expect more from the systems that we call ``artificially intelligent''. In particular, we expect them to be able to answer questions whose answers are not explicitly stated in the text, but can be easily inferred from the text. Consider the example in Figure~\ref{fig:baseline_example}, using cosine similarity between the query and text; the retriever latches onto the text which most explicitly describes commentary on the artist's work and ignores the text which contains the answer but does not contain words like ``regarded''. This is a common failure mode for RAG systems, and it demonstrates how the retriever can be a hindrance to what is otherwise a powerful model for natural language understanding.

\begin{figure}
    \setlength{\arrayrulewidth}{0.3mm}
    \centering
    \def\arraystretch{2}
    \colorbox{offwhite}{
        \begin{tabular}{l p{10cm}}
            \textbf{Question} & All of historians speak highly of Picardo's work, is this true? Why? \\
            \arrayrulecolor{lightgray} \cdashline{1-2}[1.5pt/3pt] \arrayrulecolor{black}
            \textbf{Target Text} & ``\ldots \highlight{correct}{was somewhat frowned upon} in the 1960s and 1970s, and over half a century later is seen by archeologists and historians as a matter of significant \highlight{correct}{controversy and regret}.'' \\
            \textbf{Ground Truth} & False, because some people believe that Parrado destroyed the part of historical and architectural [sic]. \\
            \arrayrulecolor{lightgray} \cdashline{1-2}[1.5pt/3pt] \arrayrulecolor{black}
            \scriptsize{RAG} & \\
            \textbf{Retrieved Text} & ``\ldots Picardo's published architectural drawings were highly regarded. They were described as \highlight{incorrect}{``magnificent''} by the leading Spanish restoration architect \ldots'' \\
            \textbf{Prediction} \raisebox{0mm} & Yes, because his architectural drawings were described as \highlight{incorrect}{``magnificent''} \ldots \\
            \arrayrulecolor{lightgray} \cdashline{1-2}[1.5pt/3pt] \arrayrulecolor{black}
            \scriptsize{\methodacronym-S} & \\
            \textbf{Retrieved Text} & ``\ldots \highlight{correct}{was somewhat frowned upon} in the 1960s and 1970s, and over half a century later is seen by archeologists and historians as a matter of significant \highlight{correct}{controversy and regret}.'' \\
            \textbf{Prediction} \raisebox{0mm} & No, \ldots Some historians and archeologists have criticized his rehabilitation and restoration projects for being \highlight{correct}{pastiche} and for the \highlight{correct}{demolition of large parts of monumental buildings} \ldots This has led to \highlight{correct}{significant controversy and regret} over half a century later. \\
        \end{tabular}
    }

    \caption{Example of a question from the LooGLE \cite{loogle} dataset answered by a traditional RAG system and \methodacronym-S. The traditional system identifies the a text that describes how Picardo's work was regarded by one figure, but it fails to identify the more subtly worded target text which contains the answer. Using summary generation, \methodacronym-S is able to retrieve a more relevant passage and correctly answer the question.}
    \label{fig:baseline_example}
\end{figure}

\subsection{Related Work}

RAG is an extremely active area of AI research with the goal of improving LLM performance and safety by augmenting the generation context with useful cached information \cite{zhao2024retrievalaugmented}. The modern approach to RAG is generator-agnostic \cite{shi2023replug}. This is convenient as it allows for the use of the latest and greatest LLMs without the need for retraining; however, it places a much greater burden on the retriever to be effective and efficient.

In addition to the rather expensive and unportable practice of training domain-specific query encoders \cite{shi2023replug} and rerankers \cite{glass2022re2g}, simple modifications to the retrieval mechanism can make a profound impact on the system's performance. For example, \cite{chen2023proposition} demonstrated the benefits of shorter chunk sizes and more granular indexing, while \cite{liu2022llamaindex} and \cite{sarthi2024raptor} construct multi-resolution document stores which afford a balance of context and precision. Input transformation has also been shown to be an effective tool for improving dense retrieval. In this paradigm, we use an LLM to generate a ``pseudo-document'' from the query: a piece of text that looks like the document we want to retrieve but it is actually contrived \cite{gao2022precise}, serving as a template for the target document.

Data augmentation is an increasingly popular solution for improving dense retrieval. \cite{huang2023makeanaudio} uses an RAG system to retrieve audio clips using a text query. To achieve such behavior, they generate text aliases for each audio clip and index the clips according to these aliases. This approach serves as  \cite{chen2023proposition} demonstrates the utility of transforming RAG documents into more digestible, concise, and explicit forms. They propose ``propositional retrieval'', a method quickly adopted by the AI agent community \cite{langchainproposition} due to its effectiveness, portability, and runtime efficiency. Their method preprocesses documents to extract individual propositions from the text and indexes them instead of chunks of the original text. This greatly improves retrieval of information which is explicitly stated, but it sacrifices nuance that may be necessary for answering more complex questions.

\subsection{Contribution}
In light of previous attempts to improve dense retrieval, we sought to develop an extension to RAG that enables retrieval of implicit content without sacrificing nuance or portability. Specifically, we make the following contributions:
\begin{itemize}[leftmargin=15pt]
\item \methodacronym: A novel framework for RAG which retrieves more relevant context by matching the query against LLM derived features
\item Implemented multiple variants of \methodacronym to improve context retrieval in RAG systems
\item Applications of \methodacronym that outperform vanilla RAG in QA accuracy by 15\%\textsuperscript{\ref{asterisk}}
\item Applications of \methodacronym that improve retrieval of relevant context by 15\%\footnote{\label{asterisk}task dependent}
\end{itemize}

\section{Method}
\begin{figure}
    \centering
    \includegraphics[width=\linewidth]{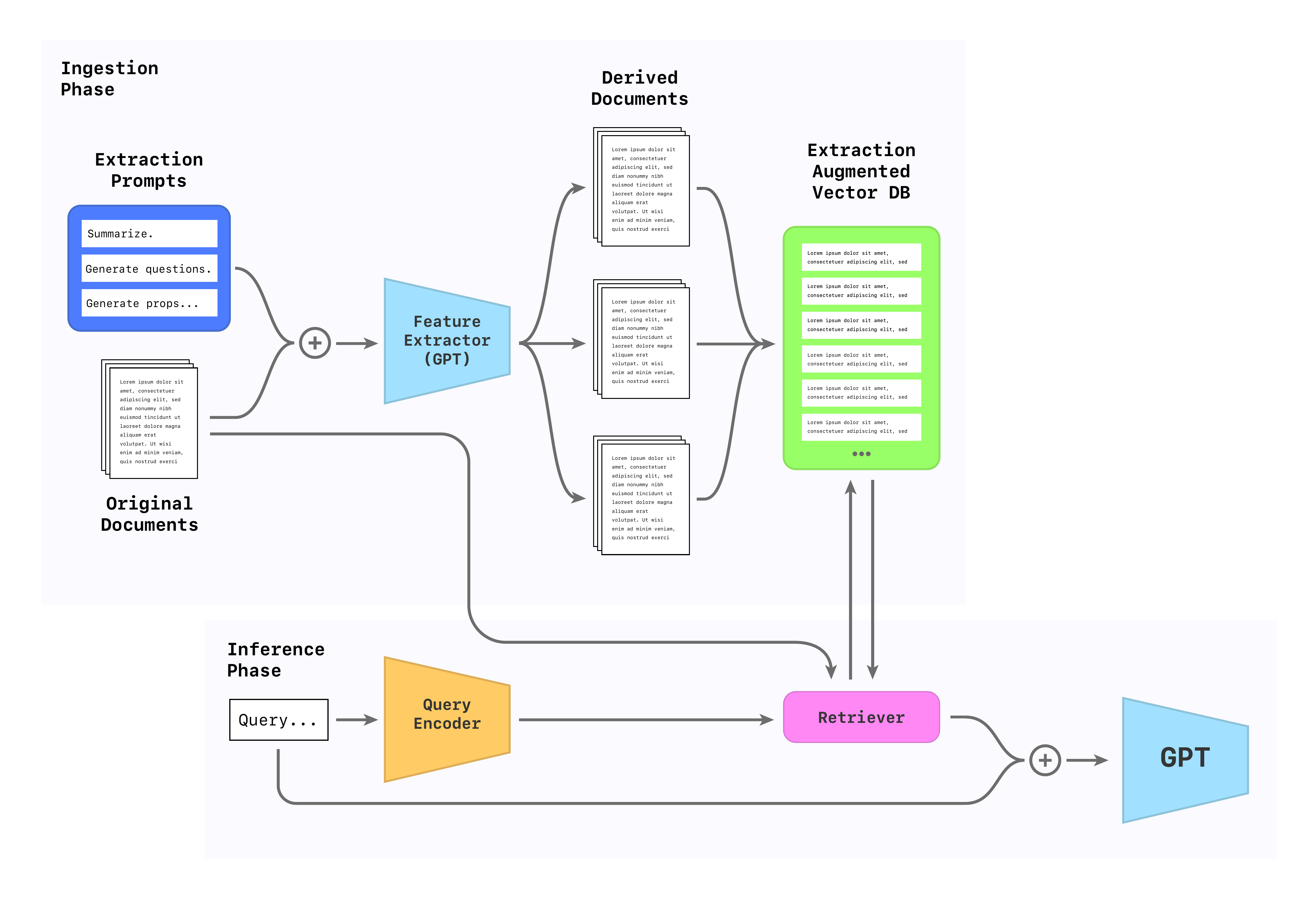}
    \caption{During the document ingest phase, \methodacronym derives new documents from the input by prompting a GPT feature extractor to summarize and generate questions from the original documents. At inference, the retriever identifies the most relevant derived documents and places the corresponding source documents into context for question answering.}
    \label{fig:diagram}
\end{figure}

\subsection{Derived Document Association}
\methodacronym generalizes the retrieval paradigms first introduced by Chen et al. (2023) \cite{chen2023proposition}. In this original paradigm, LLM generated ``propositional'' documents completely supplant the original text, throughout the retrieval and generation processes. Our method, \methodacronym, diverges from this paradigm by placing the original text in the generation context, rather than the derived documents themselves. We make this change in an effort to minimize the impact of information loss in the document derivation process, retaining the same level of detail as vanilla RAG. We call this practice of using the derived documents as handles for the original text ``derived document association''. Derived document association generalizes the retrieval technique of Huang et al. (2023), in which the system retrieves audio clips based on how well their derived text aliases match the query.

More concretely, our method involves two phases: ingest and inference. During ingest, we prompt an LLM to generate derived documents from each input document. These derived documents are either summaries of each chunk, sets of questions from each chunk, or both. We generate embeddings from these derived documents and store them in a vector database. At inference time, the retriever matches our query against all of the derived documents. For each of the top $k$ derived documents, we place the corresponding original document into our question answering context.

\subsection{\methodacronym Flavors}
\paragraph{\methodacronym-S}
Consider the following scenario where the document store has two documents: (a) ``Johnny mixed `ocean' and `fire-engine' on his palette'' and (b) ``Johnny made so much green as a world-renowned painter'', and the model is asked ``What color paint did Johnny make?''. Modern LLMs are more than capable of understanding that ``green'' is a euphemism for money, and that ``ocean'' and ``fire-engine'' are shades of blue and red that make purple when combined. However, dense retrieval with OpenAI's text-embedding-ada-002 \cite{ada002} scores document (b) 4\% higher than document (a), likely because of its inclusion of words like ``made'', ``green'', and ``paint''. While this is a contrived example, it demonstrates how dense retrieval systems place too much responsibility on a underpowered retriever, despite having access to incredibly performant natural language understanding models.

\methodacronym-S seeks to improve the retrieval of implied information by retrieving text chunks based on their LLM generated summaries. By forcing our LLM to make the document more direct, concise, and explicit, we hope to improve the retriever's ability to identify the most relevant text. We generate summaries for each chunk of text in the document store, requesting that the LLM provide concept-level summaries which paraphrase the original text. We condition this generation with the original text as well as the summary of the previous chunk to maintain coherence. \methodacronym-S+ augments this by also directly retrieving raw text chunks if they rank higher than summaries.

\paragraph{\methodacronym-S ICL}
We use the LLM as an optimizer within a feedback loop mechanism aimed at enhancing the accuracy and relevance of its responses to QA task. The LLM generates answers to a set of input questions. Each generated answer is then evaluated against a ground truth answer, and the LLM assigns itself a self-score ranging from 0 to 1, indicating the confidence or quality of its response. These scores, along with the generated answers and the respective ground truth answers, form a set of data points known as solution-score pairs.\\
These solution-score pairs are utilized as a ``meta-prompt'' for the LLM. The meta-prompt also includes a task description that contextualizes the next question for which a solution is sought. This comprehensive meta-prompt serves as a refined input that guides the LLM in adjusting its internal strategies for generating solutions, effectively using its past performance (as reflected by the scores and solution accuracy) to optimize future outputs.\\
Through this iterative process, it optimizes its context retrieval and answer generation to increase alignment with expected results.
\begin{figure}
    \centering
    \includegraphics[width=0.6\linewidth]{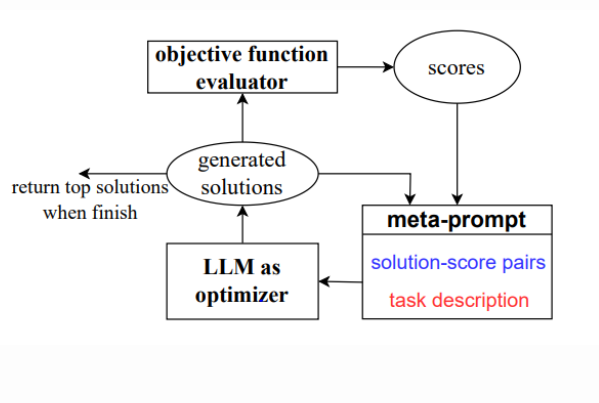}
    \caption{In-context learning using an LLM as an optimizer}
    \label{fig:diagram}
\end{figure}

\paragraph{\methodacronym-Q}
For \methodacronym-Q, we draw inspiration from the work of \cite{gao2022precise} and \cite{wang2023query}. While in their work, they generate pseudo-documents from the query in order to provide the retriever with a template for the target document, we perform the inverse process, generating pseudo-queries from the documents. We prompt the LLM to generate a set of unique reading comprehension questions from each chunk of text in the document store. To accommodate multiple chunk sizes, we generate 32 questions for every 1024 tokens of text. Since each chunk now has multiple questions associated with it, during retrieval, we retrieve the top $k$ questions that correspond to unique chunks of the original text. We then place the original text into context for question answering. Like \methodacronym-S+, \methodacronym-Q+ is a variation which also retrieves raw text chunks.

\paragraph{\methodacronym-U}
\methodacronym-U attempts to ensemble the benefits of \methodacronym-S+ and \methodacronym-Q+ by retrieving from an index which contains the union of the summary, question, and raw text document stores. \methodacronym-U emulates a form of multiheaded attention, in which each our raw text, summary, and generated questions can be thought of as multiple keys for the same value.

\subsection{Dataset and Evaluation}
We evaluate long-context question answering performance using the long-dependency QA subset of LooGLE \cite{loogle}. To minimize cost, we use the first 100 questions of the dataset, as our validation set for choosing hyperparameters, and we use the next 100 questions to test our final configurations. We compare each of our models to a standard dense retrieval baseline using the same embedding and generator models. Question answering performance is measured using ROUGE \cite{lin2004rouge} and accuracy. The accuracy of each response is decided by GPT-4; we prompt the model to decide whether the generated answer is sufficiently similar to the ground-truth, given the question.

Note that generating the summarizations and questions used as the AI derived documents takes place during document ingest and is just a means to an end that is answering the questions.

\section{Experiments}
\subsection{Implementation Details}
Our experiments are implemented using the LlamaIndex \cite{liu2022llamaindex} RAG library. For generating derived documents, we prompt an instruction tuned Mixtral-8x7B model \cite{jiang2024mixtral}. We use the frontier-class Mistral Large model \cite{mistral_large} as our question answering model. Our retriever uses a simple cosine similarity metric between the query and document embeddings, which are encoded using OpenAI's text-embedding-ada-002 \cite{ada002}.
\subsection{Results}

\begin{table}[ht]
    \centering
    \caption{Performance comparison of various flavors of \methodacronym on the validation set. }
    \label{tab:performance_comparison_val}
    \begin{tabular}{l l l l r r r}
        \toprule
        Method & Chunk Size & Overlap & Top-$k$ & Accuracy      & ROUGE-1           & ROUGE-L           \\
        \midrule
        \multirow{6}{*}{Baseline}
               & 64         & 10      & 32      & 0.43             & 0.216             & 0.174             \\
               & 128        & 25      & 16      & 0.46             & 0.209             & 0.167             \\
               & 256        & 50      & 8       & 0.48             & 0.249             & 0.206             \\
               & 512        & 100     & 4       & 0.48             & 0.223             & 0.189             \\
               & 1024       & 200     & 2       & 0.39             & 0.212             & 0.167             \\
               & 2048       & 400     & 1       & 0.35             & 0.191             & 0.152             \\
        \midrule
        \multirow{6}{*}{\methodacronym-S}
               & 64         & 10      & 32      & 0.48             & 0.214             & 0.171             \\
               & 128        & 25      & 16      & 0.41             & 0.204             & 0.167             \\
               & 256        & 50      & 8       & \underline{0.49} & 0.230             & 0.183             \\
               & 512        & 100     & 4       & 0.43             & 0.224             & 0.178             \\
               & 1024       & 200     & 2       & 0.31             & 0.205             & 0.178             \\
               & 2048       & 400     & 1       & 0.31             & 0.201             & 0.163             \\
        \midrule
        \multirow{6}{*}{\methodacronym-Q}
               & 64         & 10      & 32      & 0.46             & 0.217             & 0.173             \\
               & 128        & 25      & 16      & 0.46             & 0.222             & 0.180             \\
               & 256        & 50      & 8       & 0.44             & \textbf{0.269}    & \textbf{0.208}    \\
               & 512        & 100     & 4       & 0.46             & 0.218             & 0.180             \\
               & 1024       & 200     & 2       & 0.44             & 0.201             & 0.167             \\
               & 2048       & 400     & 1       & 0.39             & 0.203             & 0.164             \\
        \midrule
        \multirow{6}{*}{\methodacronym-U}
               & 64         & 10      & 32      & 0.47             & 0.220             & 0.181             \\
               & 128        & 25      & 16      & 0.46             & 0.214             & 0.172             \\
               & 256        & 50      & 8       & \textbf{0.52}    & \underline{0.254} & \underline{0.207} \\
               & 512        & 100     & 4       & 0.45             & 0.225             & 0.182             \\
               & 1024       & 200     & 2       & 0.46             & 0.209             & 0.171             \\
               & 2048       & 400     & 1       & 0.36             & 0.203             & 0.163             \\
        \bottomrule
    \end{tabular}
\end{table}

\begin{table}[h]
\centering
\caption{Performance comparison on test set. Having tried different configurations on the validation set (Table \ref{tab:performance_comparison_val}), we settled on a chunk size of 256, overlap of 50, and $k$ of 8 for the test set.}
\label{tab:performance_comparison_test}
\begin{tabular}{@{}lrrr@{}}
\toprule
Method   & Accuracy & ROUGE-1    & ROUGE-L   \\ \midrule
Baseline & 0.48                  & 0.19291               & 0.16160   \\
RAIDD-S  & \underline{0.50}      & \underline{0.20814}   & \underline{0.17116}   \\
RAIDD-S+ & 0.49                  & 0.20561               & 0.16612   \\
RAIDD-Q  & \textbf{0.52}         & 0.19178               & 0.16424   \\
RAIDD-Q+ & 0.46                  & \textbf{0.22677}      & \textbf{0.18130}   \\
RAIDD-U  & 0.48                  & 0.19130               & 0.15976   \\ \bottomrule
\end{tabular}
\end{table}



\begin{table}[htbp]
  \centering
  \caption{Per-Task performance comparison on the test set. CP-CCR (Correct Prediction - Correct Chunk Retrieved) and IP-CCR (Incorrect Prediction - Correct Chunk Retrieved) measures whether the retrieved chunks for both correct and incorrect predictions match the ground truth context required for answering the question. This helps to measure the performance of the methods in enhancing context retrieval. \textbf{Bold} denotes better than baseline accuracy.}
  \label{table:retrieval_perf_test}
  \begin{tabular}{l l r r r r r} 
    \toprule
    \thead{Method} & \thead{\multicolumn{1}{c}{Metric}} & \thead{Timeline \\ Reorder} & \thead{Multiple \\ Information\\ Retrieval} & \thead{Comprehension \\ and Reasoning} & \thead{Computation} & \\ 

    \midrule
    Baseline & Accuracy & 30.77\% & 40.00\% & 57.14\% & 58.82\% \\
             & CP-CCR & 75.00\% & 35.71\% & 60.00\% & 100.00\% \\
             & IP-CCR & 77.78\% & 42.86\% & 26.67\% & 42.86\% \\
    \midrule
    \methodacronym-Q & Accuracy & 30.77\% & 31.43\% & \textbf{60.00\%} & 47.06\% \\
                  & CP-CCR & 75.00\% & 18.18\% & 57.14\% & 87.50\% \\
                  & IP-CCR & 33.33\% & 41.67\% & 14.29\% & 33.33\% \\
    \midrule
    \methodacronym-Q+ & Accuracy & \textbf{38.46\%} & 37.14\% & 51.43\% & 52.94\% \\
                              & CP-CCR & 80.00\% & 46.15\% & 61.11\% & 88.89\% \\
                              & IP-CCR & 75.00\% & 31.82\% & 17.65\% & 50.00\% \\
    \midrule
    RAIDD-S ICL & Accuracy & 23.08\% &	11.43\% &	22.86\% &	5.88\% \\
                      & CP-CCR & 33.33\% &	25.00\% &	75.00\% &	100.00\% \\
                      & IP-CCR & 70.00\% & 35.48\% & 33.33\% & 75.00\% \\
    \midrule
    \methodacronym-S & Accuracy & 23.08\% & \textbf{45.71\%} & \textbf{62.86\%} & 47.06\% \\
                       & CP-CCR & 66.67\% & 43.75\% & 54.55\% & 100.00\% \\
                       & IP-CCR & 60.00\% & 26.32\% & 23.08\% & 55.56\% \\
        
    \midrule
    \methodacronym-S+ & Accuracy & \textbf{38.46\% }& \textbf{45.71\%} & \textbf{62.86\%} & 58.82\% \\
                        & CP-CCR & 80.00\% & 43.75\% & 63.64\% & 100.00\% \\
                        & IP-CCR & 62.50\% & 36.84\% & 23.08\% & 42.86\% \\
    \midrule
    \methodacronym-U & Accuracy & \textbf{46.15\%} & \textbf{42.86\%} & \textbf{62.86\%} & 52.94\% \\
              & CP-CCR & 83.33\% & 40.00\% & 50.00\% & 88.89\% \\
              & IP-CCR & 57.14\% & 35.00\% & 23.08\% & 50.00\% \\
    \bottomrule
  \end{tabular} \\
  \smallskip
   \raggedright\small
  
\end{table}

\subsection{Per-Task Analysis}
Table \ref{table:retrieval_perf_test} highlights the performance of various augmentation methods applied to RAG system across multiple tasks: timeline reorder, multiple information retrieval, comprehension and reasoning, and computation (see appendix \ref{section:question_types} for examples). These results give a clearer understanding of how each augmentation strategy impacts performance in specific domains. Here's a detailed breakdown:

\subsubsection{Timeline Reorder}
Our baseline method demonstrated a relatively modest performance, achieving an accuracy of 30.77\%, which suggests a limited capability in the timeline reordering question type. The RAIDD-U method, however, showed a significant improvement, with 46.15\% accuracy, a 15\% increase compared to the baseline. This enhancement is also reflected in the correct chunk match rate, indicating that our method is effective in retrieving more relevant context to accurately answer the questions.

Additionally, the RAIDD-Q+ and RAIDD-S+ methods have outperformed the baseline in terms of both accuracy and context retrieval, affirming their efficacy in handling timeline reorder questions. It is noteworthy that the RAIDD-S method exhibited a marked decrease in performance. We hypothesize that this decline is due to the nature of the timeline reorder task, which requires precise and factual information from the text to correctly analyze and reorder events. The summarization process inherent to the RAIDD-S method potentially omits crucial details, leading to a loss of essential information and consequently, lower performance on this task type.

\subsubsection{Multiple Information Retrieval} 
The RAIDD-S+ method demonstrates a notable improvement in performance compared to the baseline. Similarly, the RAIDD-S and RAIDD-U methods also contribute to enhanced performance, indicating that these approaches effectively enhance the retrieval of relevant context necessary for answering the questions.

However, it is observed that the question-based feature generation methods lead to a decline in performance. This decline can be attributed to the specific requirements of this question type, which likely necessitates long-range context dependencies that span multiple text chunks. The questions generated from each individual chunk do not adequately reflect the types of questions typically associated with this question type. As a result, these methods fail to retrieve the relevant context needed to accurately answer the questions.
\subsubsection{Comprehension and Reasoning}
We observe that all methods, with the exception of \methodacronym-S ICL, show enhanced performance compared to the baseline. Notably, despite its lower accuracy, the \methodacronym-S ICL method demonstrates superior capability in retrieving context pertinent to the questions posed. There is an evident 15\% increase in the extraction of relevant chunks. This suggests that providing additional context in the prompt facilitates more effective retrieval of pertinent information. However, we hypothesize that the utilization of significantly larger prompts might lead to a dilution of information. Consequently, the LLM struggles to concentrate on the retrieved context to accurately answer the question, thereby impairing performance.Despite this, we observe a consistent performance gain using the RAIDD methodology for this particular question type.
\subsubsection{Computation}
In this evaluation, it is observed that all the methods under consideration perform suboptimally in comparison to the baseline. Intriguingly, despite the baseline's capacity to retrieve relevant chunks with 100\% accuracy, it achieves only about 58\% overall accuracy. This outcome suggests that for this specific question type, the efficacy of the LLM in generating outputs is paramount. The LLM's ability to intricately process details and synthesize responses based on the provided context is crucial.

Further analysis indicates that enhancing the vector database to improve context retrieval does not significantly impact the performance for this type of question. The fundamental challenge appears to be the LLM’s capability to effectively utilize the context to compute and generate precise answers. This underscores the necessity of focusing on improving the LLM's generative abilities rather than merely augmenting the context retrieval mechanisms.

\section{Conclusion}
The results clearly demonstrate the significant impact that augmentation strategies—such as question-focused approaches, raw text integration, and summarization techniques—have on enhancing the performance of RAG systems across diverse complex tasks. The effectiveness of these strategies varies with the nature of the task, highlighting the importance for a tailored system design that incorporates specific augmentation methods suitable to the task requirements. For example, tasks requiring deep comprehension and information synthesis are better served by summarization techniques that distill and captures key details, while those requiring precise fact retrieval benefit from improved raw text integration. This adaptability in strategy selection is crucial for optimizing RAG system performance across varying task demands.

Although our method successfully retrieves enhanced contextual information to address queries, the primary obstacle in generating accurate responses lies in the capabilities of the LLM employed to produce answers. Thus, even with improved context retrieval, there remains a critical need for a more advanced LLM capable of utilizing this context to generate responses that are conceptually and semantically correct.

RAIDD introduces a novel framework for RAG which retrieves more relevant context than vanilla RAG implementations while retaining the portability of dense retrieval. We show that across multiple QA tasks, RAIDD is able to improve retrieval and QA accuracy by up to 15\%. Further research is necessary to tune the derived document generation process to retain all necessary information across domains. Additionally, we hope to develop additional types of derived documents, beyond summaries and questions, to optimize the distillation of information performed by RAIDD.

\appendix

\bibliographystyle{plain}
\bibliography{refs}

\section{Question Examples}
\label{section:question_types}
{\setlength{\arrayrulewidth}{0.3mm}
\centering
\def\arraystretch{2}
\colorbox{offwhite}{
    \begin{tabular}{l p{10cm}}
        \textbf{Timeline Reordering} & \\
        \textbf{Question} & Picardo dedicated much of his professional life to Paradores, please order these projects by open day: 1. Parador de Arcos de la Frontera, 2. Parador de Guadalupe, 3. Parador de Carmona \\
        \textbf{Target Text} & ``Restoration at Guadalupe started in November 1963 and the hotel, with twenty double rooms, opened on 11 December 1965.'' \\
        & ``Parador de Arcos de la Frontera opened to guests on 7 November 1966.'' \\
        \textbf{Ground Truth} & 2,1,3 \\
        \arrayrulecolor{lightgray} \cdashline{1-2}[1.5pt/3pt] \arrayrulecolor{black}
        \textbf{Multiple Information Retrieval} & \\
        \textbf{Question} & How did Picardo build the Parador at the Castillo de Santa Catalina to get a good view for vistors? \\
        \textbf{Target Text} & ``Using the elongated site at the top of the hill, Picardo planned a dining room, a lounge, service accommodation and guest rooms.'' \\
        & ``He styled his new building on the layout and dimensions \ldots'' \\
        \textbf{Ground Truth} & He used the elongated site at the top of the hill, styled his new building on the layout and dimensions of the old castle and on what had been discovered during his research of its surviving interior designs. \\
        \arrayrulecolor{lightgray} \cdashline{1-2}[1.5pt/3pt] \arrayrulecolor{black}
        \textbf{Comprehension and Reasoning} & \\
        \textbf{Question} & How many religious functional zones that have historically emerged in Barcelona? \\
        \textbf{Target Text} & ``Jewish Quarter (Call) :During medieval times Barcelona had a Jewish quarter\ldots'' \\
        & ``Christian Quarters: Various population centers (vila nova) were created, generally around churches and monasteries: this was the case around the church\ldots'' \\
        \textbf{Ground Truth} & 2 \\
        \arrayrulecolor{lightgray} \cdashline{1-2}[1.5pt/3pt] \arrayrulecolor{black}
        \textbf{Computation} & \\
        \textbf{Question} & How many years did Picardo work for Parador from his first of Parador projects until the bankruptcy on Parador? \\
        \textbf{Target Text} & ``For his first of many Parador projects Picardo was appointed by the Ministry of Information and Tourism \ldots'' \\
        & ``The Pedraza Hoster\'ia continued in operation until 15 December 1992 when economic pressures on the Parador chain caused its closure.'' \\
        \textbf{Ground Truth} & Twenty-nine years. \\
    \end{tabular}
}}

\end{document}